\documentclass[10pt,twocolumn,letterpaper]{article}

\usepackage{iccv}
\usepackage{times}
\usepackage{epsfig}
\usepackage{graphicx}
\usepackage{amsmath}
\usepackage{amssymb}
\usepackage{booktabs}
\usepackage{multirow}
\usepackage{subfigure}
\usepackage{xcolor}
\usepackage{helvet}
\usepackage{lineno}
\usepackage{color}
\usepackage{enumitem}
\usepackage[ruled,vlined,linesnumbered]{algorithm2e}
\usepackage{colortbl}

\usepackage[pagebackref=true,breaklinks=true,letterpaper=true,colorlinks,bookmarks=false]{hyperref}

\iccvfinalcopy 

\def\iccvPaperID{2441} 

\ificcvfinal\fi

\begin{document}

\title{On Evolving Attention Towards Domain Adaptation}

\author{Kekai Sheng$^{1}$\quad Ke Li$^{1}$\quad Xiawu Zheng$^{2}$\quad Jian Liang$^{3}$\\ Weiming Dong$^{4}$\quad Feiyue Huang$^{1}$\quad Rongrong Ji$^{2}$\quad Xing Sun$^{1}$\thanks{Corresponding author} \\
$^1$Youtu Lab, Tencent\quad $^2$Xiamen University\quad $^3$National University of Singapore\quad $^4$NLPR, CASIA \\
{\tt\small \{saulsheng, tristanli, winfredsun, garyhuang\}@tencent.com, weiming.dong@ia.ac.cn,} \\ {\tt\small liangjian92@gmail.com, zhengxiawu@stu.xmu.edu.cn, rrji@xmu.edu.cn,}
}

\maketitle
\ificcvfinal\thispagestyle{empty}\fi

\newcommand{\iMethod}{EvoADA}

\newcommand{\SOTA}{\textit{state-of-the-art}~}

\newcommand{\Sdomain}{\mathcal{S}}
\newcommand{\Tdomain}{\mathcal{T}}
\newcommand{\Sdata}{x^{\mathcal{S}}}
\newcommand{\Slabel}{y^{\mathcal{S}}}
\newcommand{\Snum}{N^{\mathcal{S}}}
\newcommand{\Tdata}{x^{\mathcal{T}}}
\newcommand{\Tnum}{N^{\mathcal{T}}}

\newcommand{\sspace}{\mathcal{A}}

\newcommand{\seed}{\mathcal{G}_{seed}}
\newcommand{\randseed}{\mathcal{G}_{rand}}
\newcommand{\crossover}{\mathcal{G}_{crossover}}
\newcommand{\mutate}{\mathcal{G}_{mutate}}
\newcommand{\best}{\mathcal{G}_{best}}

\newcommand{\RNum}[1]{\uppercase\expandafter{\romannumeral #1\relax}}

\definecolor{mygray2}{gray}{.7}

\begin{abstract}
Towards better unsupervised domain adaptation (UDA), recently, researchers propose various domain-conditioned attention modules and make promising progresses.
However, considering that the configuration of attention, \ie, the type and the position of attention module, affects the performance significantly, it is more generalized to optimize the attention configuration automatically to be specialized for arbitrary UDA scenario.
For the first time, this paper proposes \iMethod: a novel framework to evolve the attention configuration for a given UDA task without human intervention.
In particular, we propose a novel search space containing diverse attention configurations.
Then, to evaluate the attention configurations and make search procedure UDA-oriented (transferability + discrimination), 
we apply a simple and effective evaluation strategy: 1) training the network weights on two domains with off-the-shelf domain adaptation methods; 2) evolving the attention configurations under the guide of the discriminative ability on the target domain.
Experiments on various kinds of cross-domain benchmarks, \ie, Office-31, Office-Home, CUB-Paintings, and Duke-Market-1510, reveal that the proposed \iMethod\ consistently boosts multiple \SOTA\  domain adaptation approaches, and the optimal attention configurations help them achieve better performance.
\end{abstract}

\section{Introduction}
\begin{figure}
    \centering
    \includegraphics[width=0.96\linewidth]{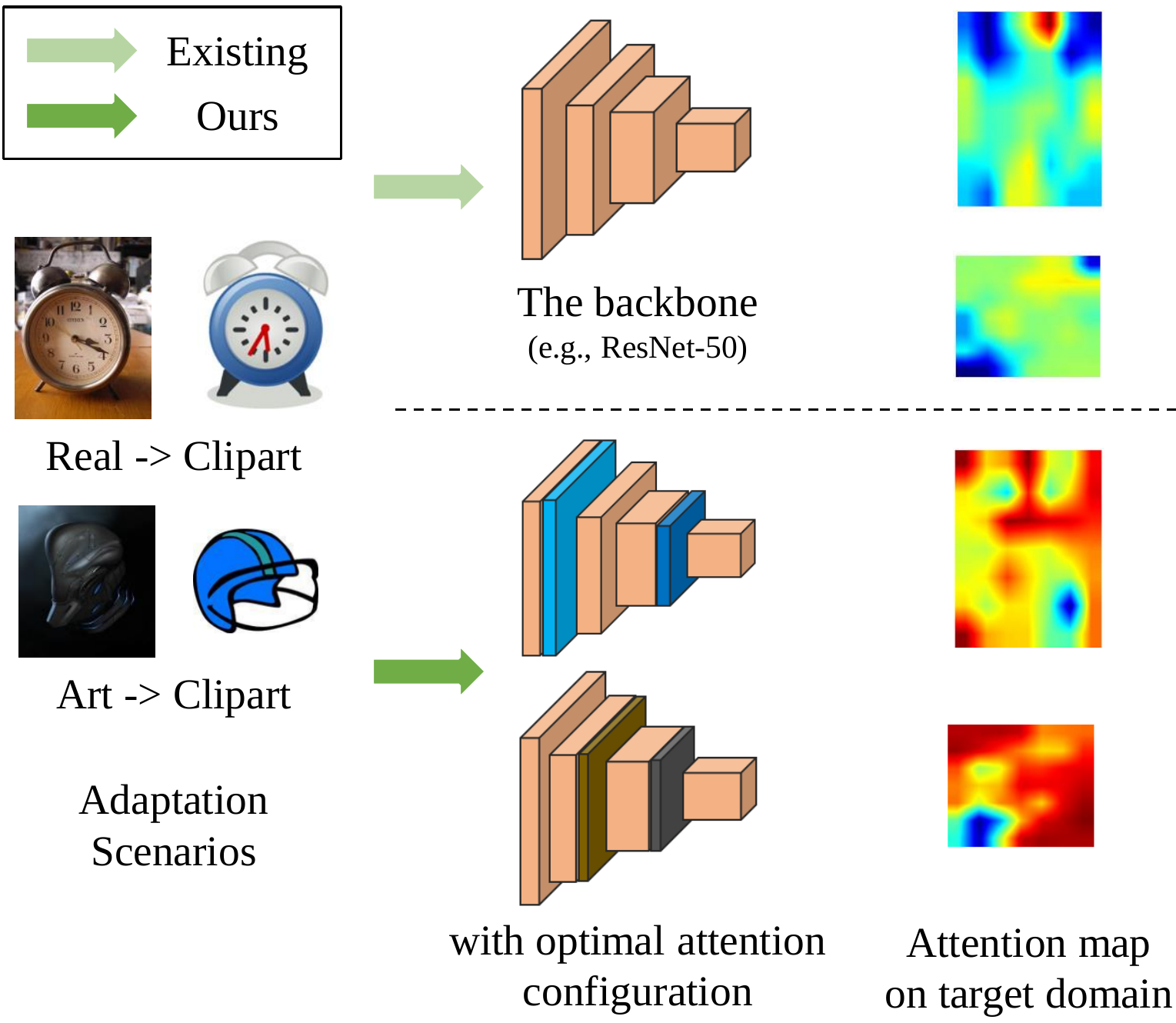}
    \caption{\textit{Why Attention Configuration Matters}. Intuitively, different configurations of attention work differently.
    Given one UDA scenario, our \iMethod\, finds which and where to add the attention modules and achieves better domain adaptation performance. 
    }
    \label{fig:Motivation}
    \vspace{-4mm}
\end{figure}

Unsupervised domain adaptation (UDA)~\cite{ganin2015DANN,long2017JAN,kang2019CAN,saito2020DANCE} aims at exploiting the meaningful knowledge from a labelled source domain to facilitate learning on another unlabelled target domain.
Generally, researchers focus on learning domain-general features. For better performance in the target domain, researchers propose domain-conditioned spatial or channel attention mechanisms~\cite{kurmi2019CADA,wang2019TN,wang2019TADA,li2020DCAN} to mitigate negative transfer and enhance the insufficient domain-specific features.
However, these works design the attention module by hand and may fall to a sub-optimal solution in real world application. For example, in Figure~\ref{fig:Motivation}, we observe that on a given UDA task and a pre-defined backbone network, different configurations of the attention module focus on various visual patterns and thus may come out with different accuracies. Therefore, given one arbitrary UDA task, a more generalized manner is to automatically find the optimal attention configuration.

One natural and widely-used solution is neural architecture search (NAS)~\cite{liu2018DARTS,zoph2018NASNet,real2019EvoNAS,guo2020SPOS}. The goal of NAS is to automatically seek for effective architectures~\cite{zoph2018NASNet,li2020adapting}.
Nevertheless, \cite{elsken2019NASASurvey,li2020adapting} point out that existing NAS algorithms rarely consider the topic of transfer learning and are vulnerable to large domain shift, resulting in inferior performance for UDA tasks.
We speculate the reasons as follows.
The first challenge is the design of search space. Existing popular search spaces (\eg, NASNet~\cite{zoph2018NASNet} or DARTS~\cite{liu2018DARTS}) are not specialized to refine the attention module and maybe ineffective in generating 
optimal architectures for domain adaptation.
The second challenge is how to evaluate the searched architectures, which is an open question. In a conventional NAS setting, we have labelled data on both the training and validation partitions, which are assumed to have little domain shift~\cite{quionero2009dataset}. It means that the architectures optimized on the training domain 
can be directly evaluated on the validation domain with ground-truth labels. But in an UDA scenario, no manual annotations are available in the target domain, 
and the relatively larger domain shift between the source and the target domains make it ineffective to evaluate the models only on the source domain. 

\begin{figure}
    \centering
    \includegraphics[width=\linewidth]{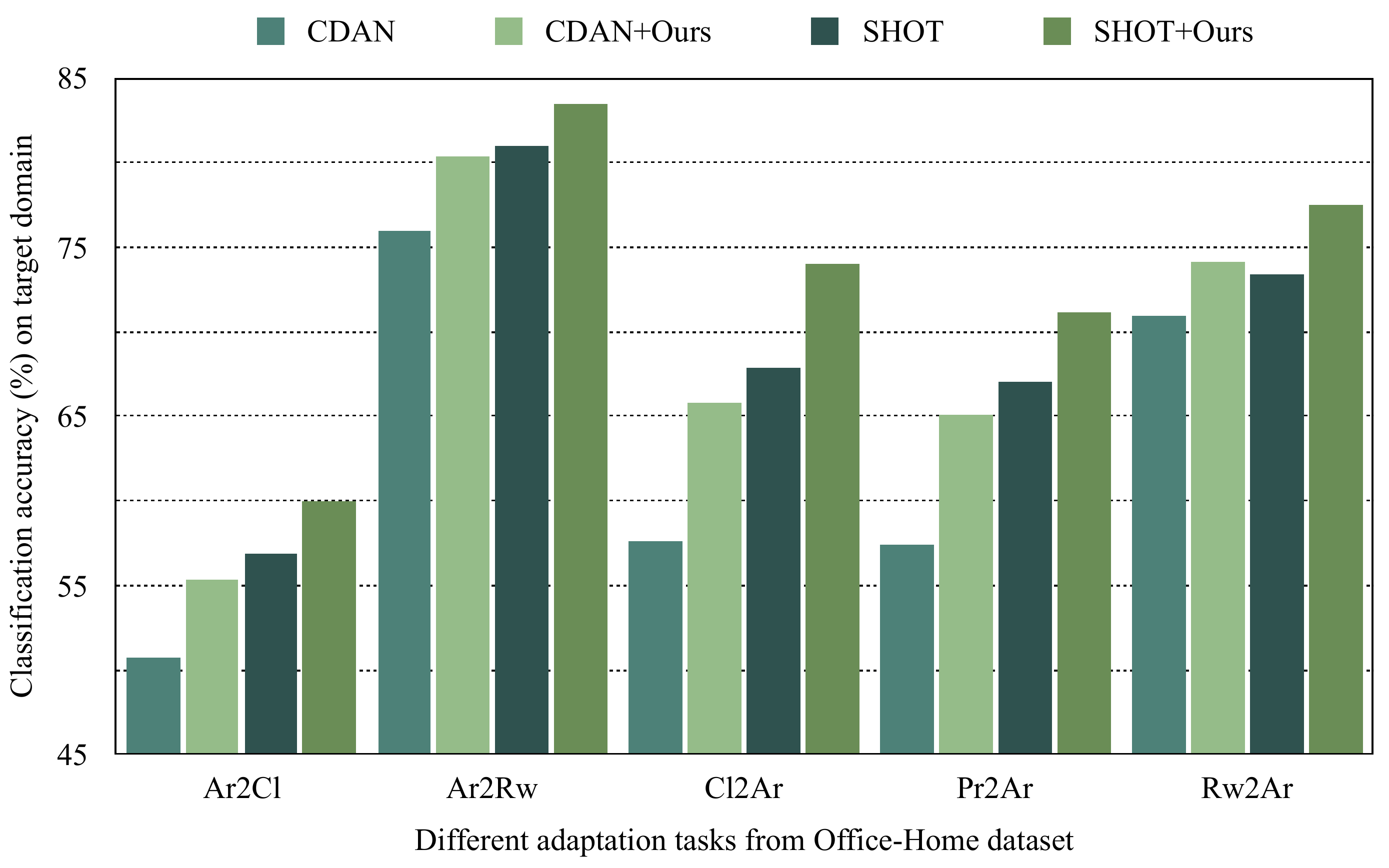}
    \caption{UDA result comparisons on Office-Home datasets.
    }
    \label{fig:SHOT_vs_SHOTandOurs}
    \vspace{-4mm}
\end{figure}

In this paper, for the first time, we propose a novel NAS algorithm, termed \iMethod.
It automatically searches attention configurations, \ie, the type and position of attention module, for different UDA scenarios.
Specifically, different from existing search spaces~\cite{zoph2018NASNet,liu2018DARTS} on the basic CNN operations, 
we design a novel search space on diverse configurations of typical attention modules~\cite{woo2018cbam,hu2018SENet,gao2019GSoP}. 
To evaluate the attention configuration effectively and make the configuration optimization UDA-oriented, we further propose a simple yet effective evaluation strategy: train the network weights on both the two domains with arbitrary UDA methods to learn transferable knowledge and then evaluate the attention configurations by our pseudo-labeling discrimination in target domain to check how the learned knowledge are discriminative on target domain. We find that the estimated qualities are strongly correlates with the final accuracy in the target domain.
Eventually, we propose a new UDA-oriented NAS scheme based on a typical evolution-based NAS algorithm.
Building on extensive experiments on $4$ cross-domain benchmarks, we verify that the searched attention configuration via our \iMethod\ benefits multiple \SOTA methods~\cite{liang2020shot,ge2020MMT,wang2020PAN} and lead to better performance in the target domain (Figure~\ref{fig:SHOT_vs_SHOTandOurs}). The experiments also provide practical insights for further research.

The main contributions are summarized as follows:
\begin{itemize}[leftmargin=*]
    \setlength{\itemsep}{0pt}
    \setlength{\parsep}{0pt}
    \item We propose a new search space with a set of effective attention modules to cover diverse attention configurations and reinforce representations for UDA tasks.
    
    \item We propose a simple yet effective strategy to evaluate the UDA performance of attention configurations. Empirically, the measure of pseudo-labeling in the target domain is effective to seek for optimal attention configurations.
    
    \item Experiments on four benchmarks verify that our algorithm successfully consolidates various \SOTA\ methods and largely promote their performance.
\end{itemize}
\section{Related Work}
\paragraph{Unsupervised Domain Adaptation} (UDA)~\cite{ganin2015DANN,long2017JAN,chen2019BSP} aims to facilitate the learning on one unlabelled target domain with the knowledge from one labelled source domain, and has practical value in many tasks~\cite{ge2020MMT,wang2020PAN,chen2021DGFaceAntiSpoofing}.
There are three typical settings: closed-set UDA, partial-set UDA (PDA), and open-set UDA (ODA). 
Technically, two fundamental problems lie in the core of UDA: 1) how to diminish the domain discrepancy on representation spaces of two domains; 2) how to deal with negative transfer and promote discrimination on target domains.
To solve the problems, researchers propose different methods that can be divided into three mainstreams: 1) feature disentanglement methods~\cite{wang2019TN,li2020DCAN}; 2) domain alignment methods~\cite{saito2018MCD,long2018CDAN}; and 3) discrimination-aware methods~\cite{zou2019CRST,cui2020BNM,liang2020shot}.
More recently, novel loss designs (\eg, ~\cite{you2019UAN,saito2020DANCE} for universal domain adaptation; ~\cite{cui2020HDA} for a progressive method) and advanced network modules (\eg, ~\cite{wang2019TN} for normalization module; ~\cite{wang2020DAFD} for convolution module; ~\cite{wang2019TADA,kurmi2019CADA,li2020DCAN} for attention module) are proposed for better domain adaptation performance.
Different from TADA~\cite{wang2019TADA}, CADA~\cite{kurmi2019CADA}, and DCAN~\cite{li2020DCAN} that combines a handcrafted attention module with an elaborate loss design, we solely investigate the improvement brought by an automatically searched attention module, which may benefit most of the \SOTA UDA methods in a more flexible way. 


\vspace{-4mm}
\paragraph{Neural Architecture Search} (NAS) aims to automate architecture engineering procedure given one certain problem.
There are five NAS mainstreams: random search, Bayesian-based method~\cite{falkner2018bohb}, reinforcement-learning approach, evolutionary scheme~\cite{back1996evolutionary}, and surrogate-based framework ($\eg$, gradient-based and predictor-based).
Representative approaches include DARTS~\cite{liu2018DARTS}, P-DARTS~\cite{chen2019progressive}, AmoebaNet~\cite{real2019EvoNAS}, one-shot NAS~\cite{bender2018oneshotNAS,guo2020SPOS}, path-level NAS~\cite{cai2018PathLevelNAS}, NASNet~\cite{zoph2018NASNet}, and AdaptNAS~\cite{li2020adapting}.
Recently, NAS technique has found its value in wide tasks, such as detection~\cite{chen2019detnas}, segmentation~\cite{liu2019AutoDeepLab}, and person re-ID~\cite{quan2019AutoReID}.
\textcolor{black}{Different from the existing literature, in this paper, we investigate the possibility and an effective solution to search for optimal architectures towards better domain adaptation. 
Concurrently, we find that Li~\etal~\cite{li2020adapting} and Robbiano~\etal~\cite{robbiano2021adversarial} have investigated a similar topic: the generalization abilities of architectures cross domains.
The differences are two-fold: 1) we propose a novel search space for diverse attention configurations, which is different from the search space of AdaptNAS~\cite{li2020adapting} (akin to NASNet~\cite{zoph2018NASNet}) and ABAS~\cite{robbiano2021adversarial} (just change the architecture of the auxiliary adversarial branch); 2) we focus on an effective NAS protocol to search for optimal attention configuration towards UDA tasks.
}






\section{Methodology}
\subsection{Preliminary}
\paragraph{\textcolor{black}{Unsupervised Domain Adaptation.}}
Formally, in one UDA task, we have a labeled dataset $\{ \Sdata_{i}, \Slabel_{i} \}_{i=1}^{\Snum}$ of $\Snum$ image-annotation pairs from source domain $\Sdomain$ and an unlabeled dataset $\{ \Tdata_{i} \}_{i=1}^{\Tnum}$ of $\Tnum$ images from target domain $\Tdomain$.
Considering that the two domains $\Sdomain$ and $\Tdomain$ are semantically related, UDA aims to facilitate the learning on $\Tdomain$ by exploiting the meaningful knowledge learned from $\Sdomain$ and try to handle the challenges of large domain shift between source domain $\Sdomain$ and target domain $\Tdomain$.

\vspace{-4mm}
\paragraph{Neural Architecture Search.}
Given one certain task, its goal to search optimal network architectures automatically. Without loss of generality, we formulate the search procedure in a bi-level optimization process: 
\begin{equation}
    \begin{split}
        \alpha &= \mathop{\arg \min}_{\alpha \in \sspace} \mathcal{L}_{val} (y, \  F(x; \  \alpha, \theta^{*}(\alpha))), \\
        s.t., \ \  &\theta^{*}(\alpha) = \mathop{\arg \min}_{\theta} \mathcal{L}_{tr}(y, \  F(x; \  \alpha, \theta)),
    \end{split}
    \label{eq:VanillaNAS}
\end{equation}
where $\alpha$ is the architecture parameter, $\sspace$ denotes the search space that contains all possible architectures, $\theta$ is the network weights, and $F(; \alpha, \theta)$ is the function of neural network.
In Eq.~(\ref{eq:VanillaNAS}), we optimize $\alpha$ with one certain loss or evaluation function on validation partition $\mathcal{L}_{val}$ under the constraint that its weight parameter $\theta^{*}(\alpha)$ is optimal for another loss function on training dataset $\mathcal{L}_{tr}$.
Researchers propose several effective NAS algorithms~\cite{real2019EvoNAS,zoph2018NASNet,liu2018DARTS} to seek for optimal architectures $\alpha$ among $\sspace$.

\subsection{\iMethod}
\begin{figure}
    \centering
    \includegraphics[width=\linewidth]{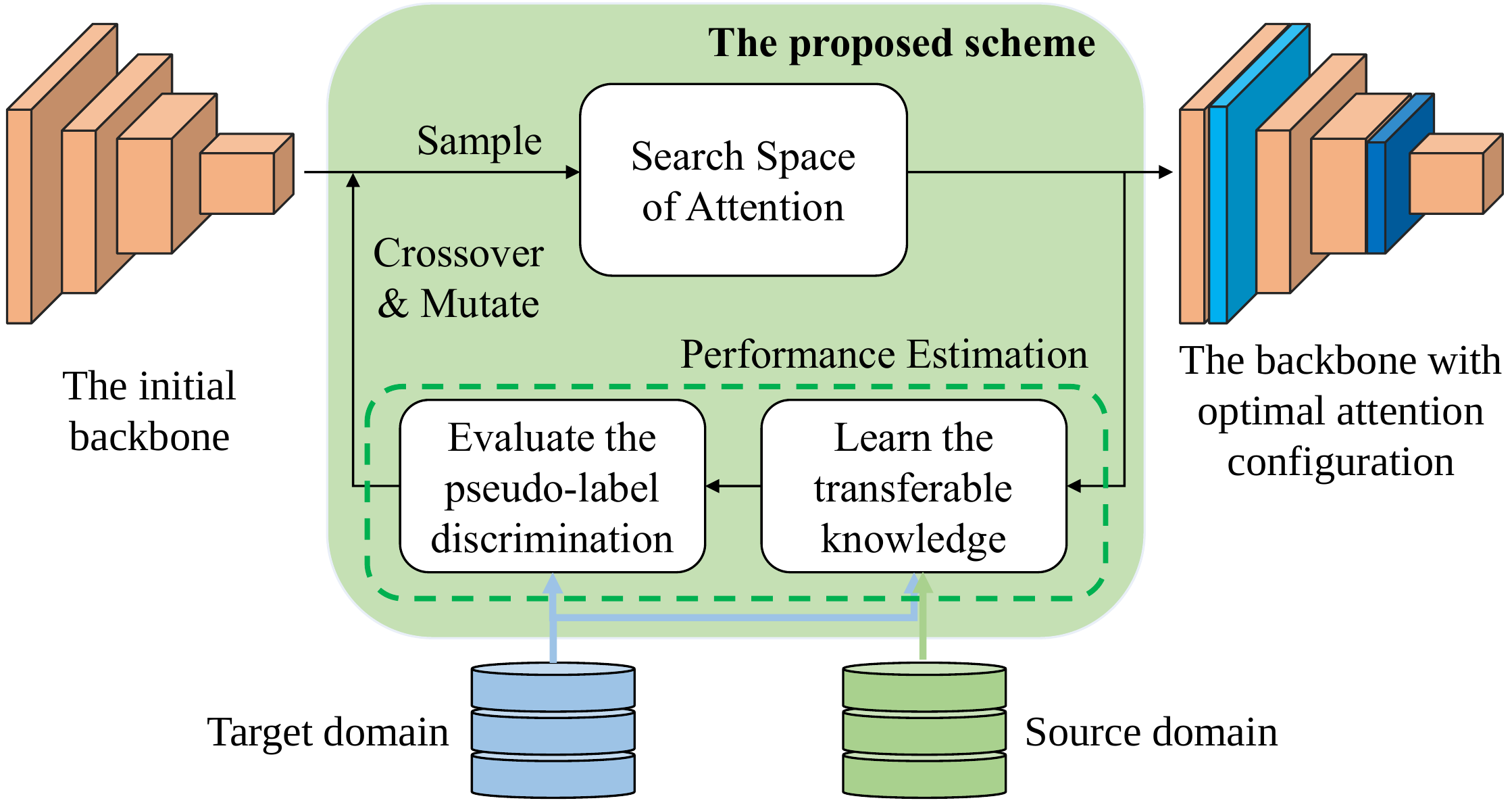}
    \caption{\textcolor{black}{The pipeline of the proposed evolutionary framework to seek for optimal attention configurations towards domain adaptation.} Given an pre-defined backbone, we sample several possible attention configurations from our search space, and then conduct an UDA-oriented performance estimation: train the network weights on the two domains with arbitrary domain adaptation method to learn transferable knowledge and evaluate the pseudo-label discrimination on target domain.}
    \label{fig:Methodology}
\end{figure}

Our goal is to investigate a more generalized UDA framework from the perspective of attention mechanism: automatically optimizing the configuration of attention in backbone networks for one arbitrary UDA scenario. To this end, we propose a new UDA-oriented NAS scheme, termed \iMethod, which searches the optimal attention in the attention configuration search space by employing our search algorithm and evaluation method. The overall illustration of the proposed method is shown in Figure~\ref{fig:Methodology}.

\begin{table}
    \centering
    \caption{The basic element to make up various attention modules. The element can also be easily extended with other kinds of attention operations for practical usages.}
    \label{tab:TypicalAttention}
    \small{
    \setlength{\tabcolsep}{3.2mm}{
    \begin{tabular}{lccc}
        \toprule
        \multicolumn{4}{c}{\textbf{Attention Module}} \\
        \hline
        \multirow{2}{*}{Module} & \multicolumn{2}{c}{Attention on} & Parameter \\
         & \textcolor{black}{spatial position} & \textcolor{black}{channel} & Choices \\
        \hline
        SE~\cite{hu2018SENet} &  & $\surd$ & \#channel \\
        GSoP~\cite{gao2019GSoP} & $\surd$ & $\surd$ & \#channel \\
        CBAM~\cite{woo2018cbam} & $\surd$ & $\surd$ & \#channel \\
        Identity &   &   & 1 \\
        \hline
        \hline
        \multicolumn{4}{c}{\textbf{Group Strategy}} \\
        \hline
        Group~\cite{wang2019fully} &  & $\surd$ & \#group \\
        \bottomrule
    \end{tabular}
    }
    }
    \vspace{-4mm}
\end{table}

\vspace{-3mm}
\paragraph{Search Space for Diverse Attention Configuration.}
Inspired by recent domain-conditioned attention mechanisms in UDA tasks~\cite{wang2019TN,li2020DCAN,wang2020DAFD}, we propose a novel and effective search space that consists of diverse attention configurations, we define our search space in two aspects: the type and the position of attention module.

$\triangleright$ \textit{Type.}
Based on the function, there are two basic types of attention: spatial type and channel type, where spatial attention aims to exploit spatial interdependence and the channel one aims to make use of channel interdependence. When it comes to the choice of its parameter, there are two extra hyper-parameters, \#channel and \#group, where the \#channel and the \#group denote the number of channels and groups of intermediate features within attention layers. To produce a diverse search space towards UDA problems, we summarize the types and parameter choices of the widely-used attention modules in Table~\ref{tab:TypicalAttention} and make up various types of attention modules with these basic elements. On sampling one type of attention, the procedure is: i) choose among $4$ kinds of basic types (\texttt{SE}~\cite{hu2018SENet}, \texttt{GSoP}~\cite{gao2019GSoP}, and \texttt{CBAM}~\cite{woo2018cbam}) or \texttt{Identity}; ii) when we select the first $3$ modules, we have two additional choices to make: \#channel and \#group. In our implementation, \#channel is $4$ (\{$256$, $512$, $1024$, $2048$\}) and \#group is $4$ (\{$1$, $2$, $4$, $8$\}).
Thus, the number of possible attention modules is $(3 \times 4 \times 4 + 1)$.
It should be noticed that the diversity and completeness of Table~\ref{tab:TypicalAttention} works well to produce optimal attention configurations for UDA scenarios, as validated later in Section~\ref{sec:exp}.


$\triangleright$ \textit{Position.}
We also need to consider the \emph{position} in a backbone network to introduce attention modules, since features from different intermediate layers have different transferability~\cite{yosinski2014transferable}. Suppose there are $L$ intermediate layers in the backbone network (\eg, $L = 50$ in ResNet-50~\cite{he2016resnet}), we then have $L$ possible position choices.
Given that low-level features generally have better transferability, we choose the deeper $L / 2$ layers to cut-off unnecessary attempts. For the shallower $L / 2$ layers, we apply weight sharing strategy on the backbone, akin to one-shot NAS algorithms~\cite{bender2018oneshotNAS,guo2020SPOS}.
To further simplify the search space, on each intermediate layer within the backbone, we only select one attention module, instead of applying parallel block design.

$\triangleright$ \textit{Overall complexity.}
Put the type and the position together, we then formulate the attention configuration parameter $\alpha$ as: $\alpha = [ \alpha_{1}, \, \alpha_{2}, \, \cdots \, \alpha_{L/2} ]$. $\alpha_{i}$ indicates the attention configuration on layer $i$ of the backbone network. The overall complexity of our search space $\sspace$ is \textcolor{black}{$|\alpha_{i}|^{L/2} = (3 \times 4 \times 4 + 1)^{25} = 49^{25} \approx 1 \times 10^{42}$}.

\vspace{-3mm}
\paragraph{UDA-oriented Evaluation Strategy.}

Considering that the core of the UDA task (transferability and discrimination), we train the network weights with a certain attention configuration $\theta(\alpha)$ on the two domains to learn transferable representation and evaluate the discrimination of the models on target domain, which in our case is mainly determined by the attention configuration $\alpha$.
Formally, we re-write the typical NAS formulation Eq.~(\ref{eq:VanillaNAS}) as follows:
\begin{equation}
    \centering
    \begin{split}
        \alpha &= \mathop{\arg \min}_{\alpha \in \sspace} \cdot \mathcal{L}^{PE}_{\Tdomain} (F(x; \  \alpha, \theta^{*}(\alpha))), \\
        s.t., \ \  &\theta^{*}(\alpha) = \mathop{\arg \min}_{\theta} \mathcal{L}^{DA}_{\Sdomain + \Tdomain}(y, \  F(x; \  \alpha, \theta)), \\
    \end{split}
    \label{eq:NAS4UDA}
\end{equation}
where $\mathcal{L}^{DA}_{\Sdomain + \Tdomain}$ denotes one certain loss function for the domain adaptation task from $\Sdomain$ to $\Tdomain$ (\eg, CDAN~\cite{long2018CDAN} or SHOT~\cite{liang2020shot}); $\mathcal{L}^{PE}_{\Tdomain}$ is the evaluation function to measure the \textit{discrimination} of models in target domain and based on the information maximization loss~\cite{gomes2010discriminative,hu2017learning}, we propose a \textcolor{black}{pseudo-label discrimination} term to measure the desirable properties of ideal representations on target domain:
\begin{equation}
    \begin{split}
        \mathcal{L}^{PE}_{\Tdomain} &= \mathcal{L}^{ent}_{\Tdomain} + \mathcal{L}^{div}_{\Tdomain} + \mathcal{L}^{pse}_{\Tdomain}, \\
    \end{split}
\end{equation}
where $\mathcal{L}^{ent}_{\Tdomain}$ is the information entropy of each output prediction on target domain, $\mathcal{L}^{div}_{\Tdomain}$ is to measure the diversity of output predictions on target domain (\eg, the negative information entropy of the average output predictions), and $\mathcal{L}^{pse}_{\Tdomain}$ is the cross-entropy based on pseudo-labels $\hat{y}$ via existing self-training methods (\eg, DeepCluster~\cite{caron2018DeepCluster}). For the detailed implementation, please refer to the supplementary.

\begin{algorithm}[t]
    \small
    {
    \SetAlgoLined
    \KwData{$\{ \Sdata_{i}, \Slabel_{i} \}_{i = 1}^{\Snum}$, $\{ \Tdata_{i} \}_{i = 1}^{\Tnum}$, population size $K$}
    \KwResult{A list of optimal architectures and corresponding network weights for the DA task from $\Sdomain$ to $\Tdomain$}
    Seed Initialization: sample K seeds from the search space $\mathcal{A}$ as $\seed$, initialize their weights, and assign various seed numbers for each seed\;
    $\best$ = [], t = 1\;
    \While{t $\leq$ T}{
        train $\seed$\;
        inference the performance of $\seed$\;
        crossover top seeds to get $\crossover$\;
        mutate the bottom seeds to get $\mutate$\;
        record mature seeds in $\crossover$\ with $\best$\;
        pop poor seeds from $\mutate$\;
        
        $\seed$ = $\crossover$ + $\mutate$;\
        
        \If{$|\seed|<$ K}{
        initialize another $(K - |\seed|)$ seeds $\randseed$\;
        $\seed$ = $\seed$ + $\randseed$\;
        }
        
        t += 1\;
    }
    return $\best$\;
    \caption{\iMethod}
    \label{alg:NAS4DA}
    }
\end{algorithm}


\vspace{-3mm}
\paragraph{Overall Search Algorithm.}
To make the architecture optimization effective, we integrate the search space and the performance evaluation with an evolutionary algorithm~\cite{back1996evolutionary} and propose our \iMethod\  (elaborated in Algorithm~\ref{alg:NAS4DA}):
\begin{itemize}[leftmargin=*]
    \setlength{\itemsep}{0pt}
    \setlength{\parsep}{0pt}
    \item \textit{Sample Seed.}
    We sample $K$ possible attention configurations of $\alpha$ as initial seeds, initialize weights of the backbone on ImageNet, randomly initialize the weights of introduced attention modules, and assign each seed with different random seed numbers (\eg, \texttt{random(seed)}) to reduce uncertainty in training procedure.

    \item \textit{Inference.}
    With initialized seeds, we then run several training epochs in each population in a parallel way, update their network weights, and evaluate their UDA performance $\mathcal{L}^{PE}_{\Tdomain}$ in the target domain.

    \item \textit{Crossover and Mutation.} We conduct crossover to get even better performance in the next generation. On those poor ones, we perform two mutation strategies to explore for better seeds: 1) drop the seed and initialize another one randomly; 2) change $\alpha$ by introducing another attention module or shifting to different layers.

    \item \textit{Update and Early Stop.} To promote training efficiency and mitigate the discord between the estimated performance and the final UDA performance, we adopt several early-stopping criteria: 
    i) when the accuracy in source domain is higher than $tr_{acc}$ (\eg., $0.95$), it generally indicates the model may suffer from negative transfer;
    ii) when the seed finishes $T$ evolution iterations;
    iii) when pseudo-label accuracy in target domain keeps being poor for over $T_{d}$ (\eg, $5$ in our implementation) iterations.
\end{itemize}

We end up with several populations that contain optimal attention configurations.
To report the final performance, we re-train the optimal architectures on the two domains from scratch with one certain domain adaptation approaches.
It should be noticed that we can also apply other architecture search algorithms, such as neural network-based reinforcement learning NAS~\cite{zoph2018NASNet}.
Thanks to the flexibility of evolution-based NAS scheme, the proposed \iMethod\  can be easily applied on arbitrary domain adaptation methods and is compatible with two typical training modes of existing UDA methods: 1) single-stage mode trains on two domains simultaneously (\eg, ~\cite{long2018CDAN,wang2020PAN}); 2) two-stage mode trains first in the source domain and then in the target domain (\eg, ~\cite{liang2020shot,ge2020MMT}).

\vspace{-3mm}
\paragraph{Differences from Concurrent Methods.}
Concurrently, Li\etal~\cite{li2020adapting} and Robbiano\etal~\cite{robbiano2021adversarial} also propose to seek for better transferable network architectures.
The differences between our \iMethod\, and them are:
1) on the design of search space: AdaptNAS~\cite{li2020adapting} adopted the search space of NASNet~\cite{zoph2018NASNet} that includes basic operations in CNNs (\eg, different pooling or convolution operations) and seeks for optimal structures in a cell perspective, ABAS~\cite{robbiano2021adversarial} only changes the structure of the auxiliary branch. In \iMethod, we design a new space to produce diverse attention configurations and apply different modules on various intermediate layers of the backbone network, which is a more generalized manner;
2) on the architecture search process: AdaptNAS adopts a gradient-based differentiable scheme~\cite{liu2018DARTS}, which might result in sub-optimal solutions; ABAS leverages the BOHB~\cite{falkner2018bohb}, a Bayesian-based hyper-parameter optimization method, which is not suitable for a high-dimensional optimization problem. \textcolor{black}{We adopt an Evolution based NAS framework~\cite{back1996evolutionary}, which is a more flexible and stable test-bed for the propose of implementation.}
Experiments in Section~\ref{sec:exp} demonstrate the effectiveness and versatility of \iMethod\, in various UDA scenarios.


\section{Experiments}
\label{sec:exp}
\subsection{Setup}
\paragraph{Implementation details.}
To check the effectiveness and versatility of our NAS algorithm, we experiment on $5$ scenarios: closed-set UDA, PDA, ODA, UDA in fine-grained classification (FGDA), and UDA in person re-ID.
Without loss of generality, we select $3$ \SOTA  methods as baselines: SHOT~\cite{liang2020shot} 
for UDA, PDA, and ODA; PAN~\cite{wang2020PAN} 
for FGDA; and MMT~\cite{ge2020MMT} 
for UDA in person re-ID.
For fair comparisons,
i) we use the same backbone, \ie, ResNet-50~\cite{he2016resnet}, which is prevailing in domain adaptation literature~\cite{long2018CDAN,cui2020BNM,wang2020PAN,ge2020MMT};
2) in PDA and ODA, we follow the same data pipeline as ~\cite{cao2018SAN,cao2019ETN,liu2019STA,liang2020shot};
3) in FGDA, we use the data pipeline of PAN~\cite{wang2020PAN};
4) to reduce the uncertainty from random seeds, we train the searched architectures with baseline methods three times using different random seeds and report the average results.
On these baselines, we adopt our \iMethod\  to investigate whether the searched architectures help boost classification performance on the target domain. Experimental results also verify the versatility of our method in various domain adaptation tasks.

To train the network weights $\theta$, we use the same settings (including data augmentation, learning-rate schedule, batch-size, etc.) as the aforementioned UDA baseline methods.
\textcolor{black}{\iMethod\, run $T=100$ epochs and $K=20$ different seeds evolve.}
We implement the proposed \iMethod\, with pytorch platform~\cite{paszke2019pytorch}.
We adopt $8$ NVIDIA Tesla V-100 GPUs and it takes approximately $20$ hours to finish the search procedure on one UDA task on average.

\vspace{-4mm}
\paragraph{Datasets.} We experiment on multiple benchmarks:

\textbf{Benchmark \RNum{1}: Office-Home \& Office-31}.
\textit{Office-Home}~\cite{venkateswara2017officehome} is one challenging medium-sized dataset. It contains $12$ adaptation tasks from $4$ distinct domains: Artistic (Ar), Clip Art (Cl), Product (Pr), and Real-World (Rw).
\textit{Office31}~\cite{saenko2010office31} is a popular small-scale domain adaptation benchmark with $4,110$ images and $31$ classes. It consists of $6$ cross-domain tasks from $3$ distinct domains: Amazon (A), Webcam (W), and DSLR (D).
Follow the practice, we report classification accuracy on each adaptation task.

\textbf{Benchmark \RNum{2}: FGDA}. \textit{CUB-200-2011}~\cite{wah2011CUB-200-2011} and \textit{CUB-200-Paintings}~\cite{wang2020PAN} are datasets for fine-grained UDA.
CUB-200-2011~\cite{wah2011CUB-200-2011} is a fine-grained visual categorization dataset with $12$K bird images in $200$ species.
CUB-200-Paintings is a dataset of $3$K bird paintings collected by Wang~\etal~\cite{wang2020PAN} and its class lists are identical to CUB-200-2011. 
We follow the same data pipeline as PAN~\cite{wang2020PAN} and report classification accuracy on the two tasks.

\textbf{Benchmark \RNum{3}: UDA in person Re-ID}. \textit{Duke}~\cite{ristani2016Duke} and \textit{Market1501}~\cite{zheng2015Market1501} are two widely-used person re-ID datasets. Market-1501~\cite{zheng2015Market1501} consists of $32$K labelled images of $1,501$ identities shot from $6$ cameras. $13$K images of $751$ identities are used for training and $19.7$ images of $750$ identities are used for inference. Duke~\cite{ristani2016Duke} contains $16.5$K photos of $702$ identities for training, and photos out of additional $702$ identities for testing. 
We follow the same pipeline on these benchmarks as Ge~\etal~\cite{ge2020MMT} and report mean average precision (mAP) 
to evaluate the performance.

\begin{table*}
    \centering
    \footnotesize{
    \setlength{\tabcolsep}{1.5mm}{
    \caption{Accuracy ($\%$) on Office-Home for UDA, PDA, and ODA methods (ResNet-50).}
    \label{tab:Results_OfficeHome}
    \begin{tabular}{lccccccccccccc}
        \toprule
        \textbf{Closed-set UDA} & Ar$\to$Cl & Ar$\to$Pr & Ar$\to$Rw & Cl$\to$Ar & Cl$\to$Pr & Cl$\to$Rw & Pr$\to$Ar & Pr$\to$Cl & Pr$\to$Rw & Rw$\to$Ar & Rw$\to$Cl & Rw$\to$Pr & AVG \\
        \midrule
        ResNet-50~\cite{he2016resnet} & 34.9 & 50.0 & 58.0 & 37.4 & 41.9 & 46.2 & 38.5 & 31.2 & 60.4 & 53.9 & 41.2 & 59.9 & 46.1 \\
        DANN~\cite{ganin2015DANN} & 45.6 & 59.3 & 70.1 & 47.0 & 58.5 & 60.9 & 46.1 & 43.7 & 68.5 & 63.2 & 51.8 & 76.8 & 57.6 \\
        JAN~\cite{long2017JAN} & 45.9 & 61.2 & 68.9 & 50.4 & 59.7 & 61.0 & 45.8 & 43.4 & 70.3 & 63.9 & 52.4 & 76.8 & 58.3 \\
        CDAN~\cite{long2018CDAN} & 50.7 & 70.6 & 76.0 & 57.6 & 70.0 & 70.0 & 57.4 & 50.9 & 77.3 & 70.9 & 56.7 & 81.6 & 65.8 \\
        ABAS~\cite{robbiano2021adversarial} & 51.5 & 71.7 & 75.5 & 59.8 & 69.4 & 69.5 & 59.8 & 47.1 & 77.7 & 70.6 & 55.2 & 80.2 & 65.7 \\
        TADA~\cite{wang2019TADA} & 53.1 & 72.3 & 77.2 & 59.1 & 71.2 & 72.1 & 59.7 & 53.1 & 78.4 & 72.4 & 60.0 & 82.9 & 67.6 \\
        CADA-A~\cite{kurmi2019CADA} & 56.9 & 75.4 & 80.2 & 61.7 & 74.6 & 74.9 & 62.9 & 54.4 & 80.9 & 74.3 & \textbf{61.1} & 84.4 & 70.1 \\
        DCAN~\cite{li2020DCAN} & 54.5 & 75.7 & 81.2 & 67.4 & 74.0 & 76.3 & 67.4 & 52.7 & 80.6 & 74.1 & 59.1 & 83.5 & 70.5 \\
        SHOT~\cite{liang2020shot} & 56.9 & \textbf{78.1} & 81.0 & 67.9 & \textbf{78.4} & 78.1 & 67.0 & 54.6 & 81.8 & 73.4 & 58.1 & 84.5 & 71.6 \\
        SHOT+Ours & \textbf{60.0} & 78.0 & \textbf{83.5} & \textbf{74.0} & 77.9 & \textbf{79.8} & \textbf{71.2} & \textbf{56.3} & \textbf{82.8} & \textbf{77.5} & 59.0 & \textbf{86.2} & \textbf{73.9} \\
        \hline \hline
        \textbf{Partial-set UDA} & Ar$\to$Cl & Ar$\to$Pr & Ar$\to$Rw & Cl$\to$Ar & Cl$\to$Pr & Cl$\to$Rw & Pr$\to$Ar & Pr$\to$Cl & Pr$\to$Rw & Rw$\to$Ar & Rw$\to$Cl & Rw$\to$Pr & AVG \\
        \midrule
        ResNet-50~\cite{he2016resnet} & 46.3 & 67.5 & 75.9 & 59.1 & 59.9 & 62.7 & 58.2 & 41.8 & 74.9 & 67.4 & 48.2 & 74.2 & 61.3 \\
        DANN~\cite{ganin2015DANN} & 35.5 & 48.2 & 51.6 & 35.2 & 35.4 & 41.4 & 34.8 & 31.7 & 46.2 & 47.5 & 34.7 & 49.0 & 40.9 \\
        SAN~\cite{cao2018SAN} & 44.4 & 68.7 & 74.6 & 67.5 & 65.0 & 77.8 & 59.8 & 44.7 & 80.1 & 72.2 & 50.2 & 78.7 & 65.3 \\
        ETN~\cite{cao2019ETN} & 59.2 & 77.0 & 79.5 & 62.9 & 65.7 & 75.0 & 68.3 & 55.4 & 84.4 & 75.7 & 57.7 & 84.5 & 70.5 \\
        SAFN~\cite{xu2019SAFN} & 58.9 & 76.3 & 81.4 & 70.4 & 73.0 & 77.8 & 72.4 & 55.3 & 80.4 & 75.8 & 60.4 & 79.9 & 71.8 \\
        BA$^{3}$US~\cite{liang2020BA3US} & 60.6 & 83.2 & 88.4 & 71.8 & 72.8 & 83.4 & 75.5 & 61.6 & 86.5 & 79.3 & 62.8 & 86.1 & 76.0 \\
        SHOT~\cite{liang2020shot} & 62.8 & 84.2 & \textbf{92.3} & 75.1 & 76.3 & \textbf{86.4} & 78.5 & 62.3 & 89.6 & 80.9 & \textbf{63.8} & 87.1 & 78.3 \\
        SHOT+Ours & \textbf{66.5} & \textbf{84.7} & 89.8 & \textbf{80.3} & \textbf{80.9} & 86.3 & \textbf{83.3} & \textbf{64.1} & \textbf{90.1} & \textbf{85.5} & 61.4 & \textbf{89.9} & \textbf{80.2} \\
        \hline \hline
        \textbf{Open-set UDA} & Ar$\to$Cl & Ar$\to$Pr & Ar$\to$Rw & Cl$\to$Ar & Cl$\to$Pr & Cl$\to$Rw & Pr$\to$Ar & Pr$\to$Cl & Pr$\to$Rw & Rw$\to$Ar & Rw$\to$Cl & Rw$\to$Pr & AVG \\
        \midrule
        ResNet-50~\cite{he2016resnet} & 53.4 & 69.3 & 78.7 & 61.4 & 61.8 & 71.0 & 64.0 & 52.7 & 74.9 & 70.0 & 51.9 & 74.1 & 65.3 \\
        DANN~\cite{ganin2015DANN}     & 54.6 & 69.5 & 80.2 & 61.9 & 63.5 & 71.7 & 63.3 & 49.7 & 74.2 & 71.3 & 51.9 & 72.9 & 65.4 \\
        OSBP~\cite{saito2018OSBP}     & 56.7 & 67.5 & 80.6 & 62.5 & 65.5 & 74.7 & 64.8 & 51.5 & 71.5 & 69.3 & 49.2 & 74.0 & 65.7 \\
        STA~\cite{liu2019STA} & 58.1 & 71.6 & 85.0 & 63.4 & 69.3 & 75.8 & 65.2 & 53.1 & 80.8 & 74.9 & 54.4 & \textbf{81.9} & 69.5 \\
        ETN~\cite{cao2019ETN} & 58.2 & \textbf{79.9} & \textbf{85.5} & 67.7 & 70.9 & \textbf{79.6} & \textbf{66.2} & 54.8 & 81.2 & 76.8 & 60.7 & 81.7 & 71.9 \\
        SHOT~\cite{liang2020shot} & 60.5 & 59.2 & 69.5 & 63.4 & 73.6 & 61.8 & 54.7 & 80.4 & 81.8 & 82.3 & 82.6 & 77.2 & 70.6 \\
        SHOT+Ours & \textbf{62.1} & 60.2 & 79.2 & \textbf{69.4} & \textbf{73.6} & 63.7 & 58.1 & \textbf{82.7} & \textbf{87.0} & \textbf{87.4} & \textbf{86.5} & 79.3 & \textbf{74.1} \\
        \bottomrule
    \end{tabular}
    } 
    } 
    \vspace{-4mm}
\end{table*}
\begin{table}
    \centering
    \caption{Accuracy ($\%$) on Office-31 for UDA (ResNet-50).}
    \label{tab:Results_Office31}
    \small{
    \setlength{\tabcolsep}{1.0mm}{
    \begin{tabular}{lccccccc}
        \toprule
        Method & \scriptsize{A$\to$W} & \scriptsize{D$\to$W} & \scriptsize{W$\to$D} & \scriptsize{A$\to$D} & \scriptsize{D$\to$A} & \scriptsize{W$\to$A} & \scriptsize{AVG}\\
        \midrule
        ResNet-50~\cite{he2016resnet} & 68.4 & 96.7 & 99.3 & 68.9 & 62.5 & 60.7 & 76.1 \\
        DANN~\cite{ganin2015DANN} & 82.0 & 96.9 & 99.1 & 79.7 & 68.2 & 67.4 & 82.2 \\
        JAN~\cite{long2017JAN} & 86.0 & 96.7 & 99.7 & 85.1 & 69.2 & 70.7 & 84.6 \\
        MCD~\cite{saito2018MCD} & 88.6 & 98.5 & 100.0 & 92.2 & 69.5 & 69.7 & 86.5 \\
        CRST~\cite{zou2019CRST} & 89.4 & 98.9 & 100.0 & 88.7 & 72.6 & 70.9 & 86.8 \\
        CDAN~\cite{long2018CDAN} & 94.1 & 98.6 & 100.0 & 92.9 & 71.0 & 69.3 & 87.7 \\
        TADA~\cite{wang2019TADA} & 94.3 & 98.7 & 99.8 & 91.6 & 72.9 & 73.0 & 88.4 \\
        BSP~\cite{chen2019BSP} & 93.3 & 98.2 & 100.0 & 93.0 & 73.6 & 72.6 & 88.5 \\
        CADA-A~\cite{kurmi2019CADA} & \textbf{96.8} & \textbf{99.0} & 99.8 & 93.4 & 71.7 & 70.5 & 88.5 \\
        SHOT~\cite{liang2020shot} & 90.9 & 98.8 & 99.9 & 93.1 & 74.5 & 74.8 & 88.7 \\
        \midrule
        SHOT+Ours & 94.0 & 97.9 & 100.0 & \textbf{94.2} & \textbf{74.6} & \textbf{74.9} & \textbf{89.3} \\
        \bottomrule
    \end{tabular}
    }
    }
    \vspace{-2mm}
\end{table}

\vspace{-4mm}
\paragraph{Baselines.}
We compare with multiple $\SOTA$ approaches: DANN~\cite{ganin2015DANN}, JAN~\cite{long2017JAN}, OSBP~\cite{saito2018OSBP}, CDAN~\cite{long2018CDAN}, 
IBN-Net~\cite{pan2018IBNNet}, MCD~\cite{saito2018MCD}, SAN~\cite{cao2018SAN}, TADA~\cite{wang2019TADA}, BSP~\cite{chen2019BSP}, 
SAFN~\cite{xu2019SAFN}, STA~\cite{liu2019STA}, CADA-A~\cite{kurmi2019CADA}, 
ETN~\cite{cao2019ETN}, 
CRST~\cite{zou2019CRST}, BA$^{3}$US~\cite{liang2020BA3US}, DCAN~\cite{li2020DCAN}, MMT~\cite{ge2020MMT}, PAN~\cite{wang2020PAN}, SHOT~\cite{liang2020shot}, and ABAS~\cite{robbiano2021adversarial}. 
Among them, TADA~\cite{wang2019TADA}, CADA-A~\cite{kurmi2019CADA}, DCAN~\cite{li2020DCAN} are the competitive approaches of better attention module design towards domain adaptation and ABAS~\cite{robbiano2021adversarial} (one current work) also adopts NAS to search optimal architectures for domain adaptation, which provide a good counterpart to investigate the effect of network design in the topic of domain adaptation.

\subsection{Results on Office-Home \& Office-31}
Experiments on Office-Home benchmark in Table~\ref{tab:Results_OfficeHome} include $3$ typical settings: closed-set UDA, PDA, and ODA~\footnote{To compare with other ODA methods, we report the OS values. Results of the baseline methods come from ~\cite{liang2020shot}.}.
As we can observe that, in term of average accuracy, the proposed NAS algorithm helps SHOT achieve better performance: $+2.3\%$ on closed-set UDA tasks, $+1.9\%$ on PDA tasks, and $+3.5\%$ on ODA tasks. 
We also notice that sometimes, optimal performances can be obtained when only one GSoP~\cite{gao2019GSoP} attention layer is put at \texttt{Layer3} for ResNet-50. These observations encourage advanced development of the attention mechanism in domain adaptation problems.



Numerical results on Office-31 dataset are listed in Table~\ref{tab:Results_Office31}.
Again, the proposed \iMethod\, generally helps SHOT promote its classification performance on target domain. 
These results indicate the importance of optimal attention configuration and the effectiveness of our \iMethod\, in typical domain adaptation scenarios.

\subsection{Results on FGDA and UDA in Person Re-ID}
We also experiment on two additional cross-domain applications: FGDA tasks and UDA tasks of person re-ID.
Results are listed in Table~\ref{tab:Results_CUB200_FGDA}.
\textit{On FGDA}:
The performance gains from the searched attention configurations are generally large on both FGDA scenarios. In terms of average accuracy, our \iMethod\, helps PAN achieve $4.1\%$ gains. 
Similar to the observations previously, we find that the gains can be achieved by automatically introducing $2$ or $3$ attention modules at proper layers of the backbone network. 
\textit{On UDA in Person Re-ID}:
For full comparisons, we experiment with different configurations of MMT~\cite{ge2020MMT}: \textit{MMT-500} and \textit{MMT-700} means that in the MMT framework, $500$ and $700$ centroids are adopted when k-means clustering is used, and \textit{MMT-DBSCAN} means DBSCAN clustering is adopted for pseudo-labels.
As listed in Table~\ref{tab:Results_CUB200_FGDA}, the architecture searched by our \iMethod\  generally outperforms the other two competitive baselines, \ie, ResNet-50~\cite{he2016resnet} and IBN-Net-50~\cite{pan2018IBNNet}, over different configurations of MMT method and two UDA person re-ID task scenarios.

Together, we verify the effectiveness and versatility of the proposed NAS scheme in searching for optimal attention configurations for various domain adaptation scenarios.

\begin{table}
    \centering
    \caption{Accuracy ($\%$) on CUB-Paintings (ResNet-50) and mAP (\%) Market-1501-Duke (ResNet-50 vs IBN-Net-50 vs ours).}
    \label{tab:Results_CUB200_FGDA}
    \small{
    \setlength{\tabcolsep}{1.1mm}{
    \begin{tabular}{lccc}
        \toprule
        \multirow{2}{*}{\textbf{FGDA}} & \scriptsize{CUB-200-2011$\to$} & \scriptsize{CUB-200-Paintings} & \multirow{2}{*}{AVG} \\
         & \scriptsize{CUB-200-Paintings} & \scriptsize{$\to$CUB-200-2011} &  \\
        \hline
        ResNet-50~\cite{he2016resnet} & 47.9 & 36.6 & 42.3 \\
        DANN~\cite{ganin2015DANN} & 57.5 & 43.0 & 50.3 \\
        JAN~\cite{long2017JAN} & 62.4 & 40.4 & 51.4 \\
        MCD~\cite{saito2018MCD} & 63.4 & 43.6 & 53.5 \\
        CDAN~\cite{long2018CDAN} & 63.2 & 45.4 & 54.3 \\
        BSP~\cite{chen2019BSP} & 63.3 & 46.6 & 55.0 \\
        SAFN~\cite{xu2019SAFN} & 61.4 & 48.9 & 55.2 \\
        PAN~\cite{wang2020PAN} & 67.4 & 50.9 & 59.2 \\
        \midrule
        PAN+Ours & \textbf{70.5} & \textbf{56.0} & \textbf{63.3} \\
        \midrule
        \multirow{2}{*}{\textbf{UDA in Person ReID}} & \scriptsize{Market1501} & \scriptsize{Duke$\to$} & \multirow{2}{*}{AVG} \\
         & $\to$Duke & Market-1501 &  \\
        \midrule
        MMT-500~\cite{ge2020MMT} & 63.1 & 71.2 & 67.2 \\
        + IBN-Net-50~\cite{pan2018IBNNet} & 65.7 & 76.5 & 71.1 \\
        + Ours & 69.6 & 79.9 & 74.8 \\
        \midrule
        MMT-700~\cite{ge2020MMT} & 65.1 & 69.0 & 67.1 \\
        + IBN-Net-50~\cite{pan2018IBNNet} & 68.7 & 74.5 & 71.6 \\
        + Ours & 71.0 & 78.5 & 74.8 \\
        \midrule
        MMT-DBSCAN~\cite{ge2020MMT} & 64.3 & 75.6 & 70.0 \\
        + IBN-Net-50~\cite{pan2018IBNNet} & 68.8 & 80.5 & 74.7 \\
        + Ours & \textbf{71.4} & \textbf{84.3} & \textbf{77.9} \\
        \bottomrule
    \end{tabular}
    }
    }
    \vspace{-2mm}
\end{table}
\subsection{Ablation Study \& Insight Analysis}
\label{seq:ablation}

\begin{table}
    \centering
    \caption{Comparison of the proposed search space and two existing typical ones. The Experiments are conducted on the four closed-set UDA settings on Office-Home dataset.}
    \label{tab:VS_NASNet_DARTS_search_space}
    \small{
    \setlength{\tabcolsep}{1.8mm}{
    \begin{tabular}{lcccc}
        \toprule
        Settings & NASNet~\cite{zoph2018NASNet} & DARTS~\cite{liu2018DARTS} & ABAS~\cite{robbiano2021adversarial} & Ours \\
        \midrule
        Ar $\to$ Cl & 57.1 & 56.8 & 51.5 & \textbf{60.0} \\
        Cl $\to$ Pr & \textbf{78.1} & 77.3 & 69.4 & 77.9 \\
        Pr $\to$ Rw & 81.3 & 80.7 & 77.7 & \textbf{82.8} \\
        Rw $\to$ Ar & 73.0 & 74.6 & 70.6 & \textbf{77.5} \\
        \midrule
        AVG & 72.6 & 72.3 & 67.3 & \textbf{74.6} \\
        \bottomrule
    \end{tabular}
    }
    }
    \vspace{-2mm}
\end{table}

\paragraph{Comparison with Other Search Spaces.}
To demonstrate the effectiveness of the proposed search space in the topic of domain adaptation, we also compare with two typical search spaces in NAS methods: the search space of NASNet~\cite{zoph2018NASNet} and that of DARTS~\cite{liu2018DARTS}. Both of them are based on basic operations in convolutional neural networks (\eg, dilated convolution, pooling, and skip connection). We randomly select $4$ close-set UDA settings from Office-Home dataset and alternate our search space with the two to investigate how their performance in the context of domain adaptation.
For full comparison, we also report the results from ABAS~\cite{robbiano2021adversarial}.
As listed in Table~\ref{tab:VS_NASNet_DARTS_search_space} the proposed attention-based search space does outperform other existing alternatives and yields the best domain adaptation results.

\begin{figure}
    \centering
    \includegraphics[width=\linewidth]{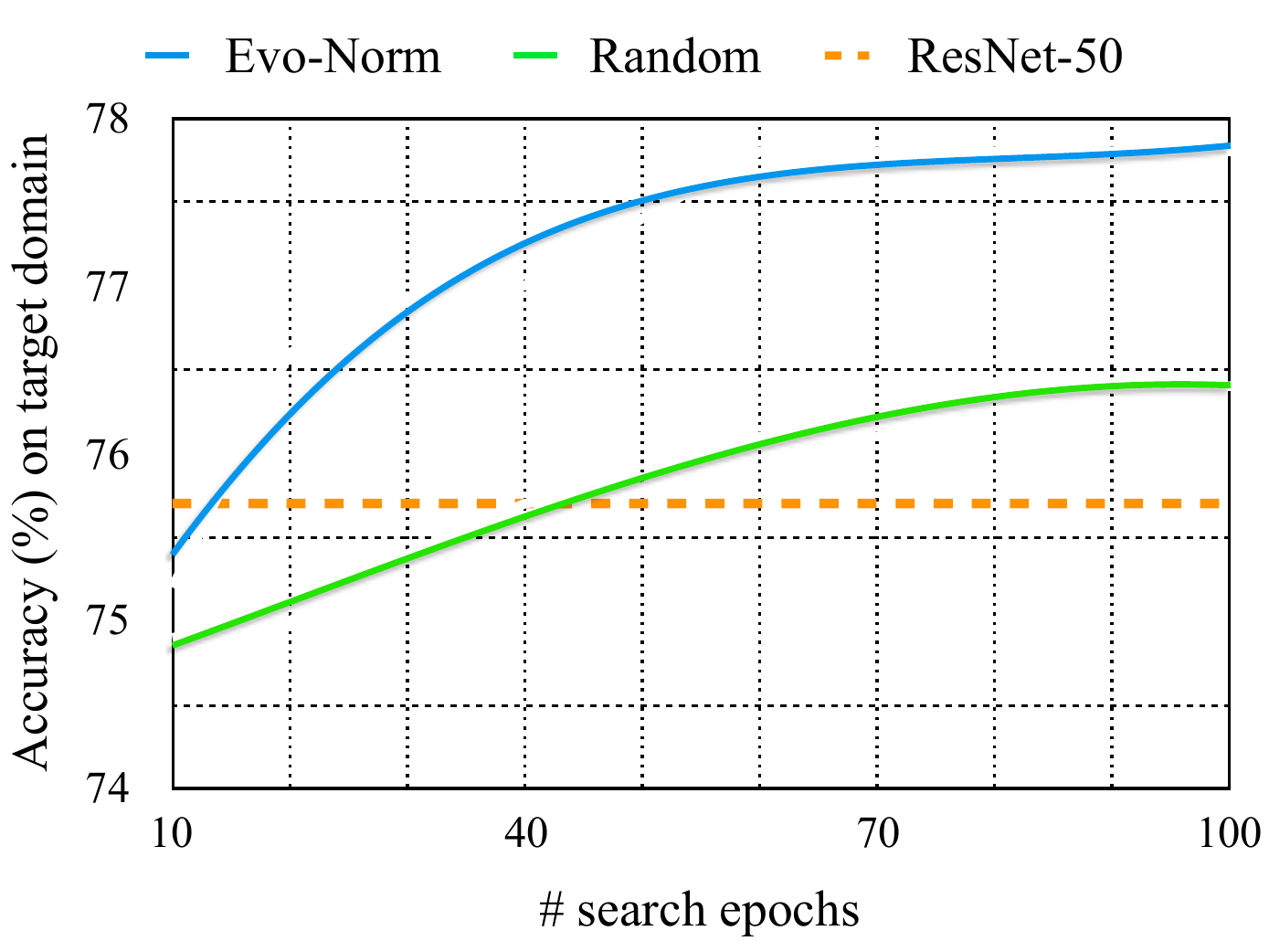}
    \caption{The comparison of \iMethod\, and random search on the partial UDA task Pr$\to$Rw on Office-Home dataset. The prevailing backbone, ResNet-50, is denoted as the dashed horizontal bar.
    }
    \label{fig:VisSearchProcess}
    \vspace{-2mm}
\end{figure}

\vspace{-2mm}
\paragraph{Comparison with Random Search.}
The search curves of our \iMethod\, and one random search algorithm are shown in Figure~\ref{fig:VisSearchProcess}.
As we can observe that our \iMethod\, is more effective in optimizing the attention configurations for domain adaptation settings.
We find similar observations on Office-Home benchmark with two additional baseline methods (random search v.s. ours): CDAN~\cite{long2018CDAN} (66.4\% v.s. 69.8\%) and SHOT~\cite{liang2020shot} (71.9\% v.s. 73.9\%).
Therefore, the results of random search demonstrate the necessity of an effective NAS algorithm towards domain adaptation tasks.

\vspace{-2mm}
\paragraph{Hyper-parameter Sensitivity.}
We investigate the sensitivity to $3$ hyper-parameters, $tr_{acc}$, $T$, $T_{d}$. Empirically, we observe that: When $t_{acc} = 0.98$, the results go worse; when $t_{acc} \in \{0.9, 0.93, 0.95\}$, the results are similar; when $t_{acc} \in \{0.8, 0.85\}$, the results become worse again. ii) When $T \ge 100$, the results are slightly better but the cost also arises. iii) When $T_d > 5$, the results are relatively worse; when $T_d \le 5$, the results are similar.
Thus, our \iMethod~ is relatively robust to these hyper-parameters.



\vspace{-2mm}
\paragraph{Effectiveness of Performance Estimation.}
To further demonstrate the rationale of our performance estimation strategy, we show the rank correlation, \ie, Spearman $\rho$, between our estimation results and the final accuracy in the target domain. For comparison, we also show the rank correlation between the accuracy in the source domain and that in the target domain. The results are shown in Table~\ref{tab:RankCorrelationComparison}.
Obvious, the rank correlation between the accuracy in the source domain and that in the target domain is relatively low, due to the large domain shift between two domains. The estimation results via our evaluation strategy, on the other hand, are highly correlated with the accuracy in the target domain and are effective to guide search procedures to seek optimal architectures for domain adaptation.

\begin{table}
    \centering
    \caption{Comparison of the rank correlation between the estimation results via the accuracy on source domain and that via our evaluation protocol.}
    \label{tab:RankCorrelationComparison}
    \small{
    \setlength{\tabcolsep}{2.4mm}{
    \begin{tabular}{lcc}
        \toprule
        Criteria & Office-31 & Office-Home \\
        \midrule
        Accuracy on Source Domain & 0.40 & 0.23 \\
        Our estimation protocol & 0.68 & 0.54 \\
        \bottomrule
    \end{tabular}
    }
    }
\end{table}





\begin{figure}
    \centering
    \subfigure[]{
        \label{fig:ResultsOf500Configurations}
        \begin{minipage}{\linewidth}
        \includegraphics[width=1\linewidth]{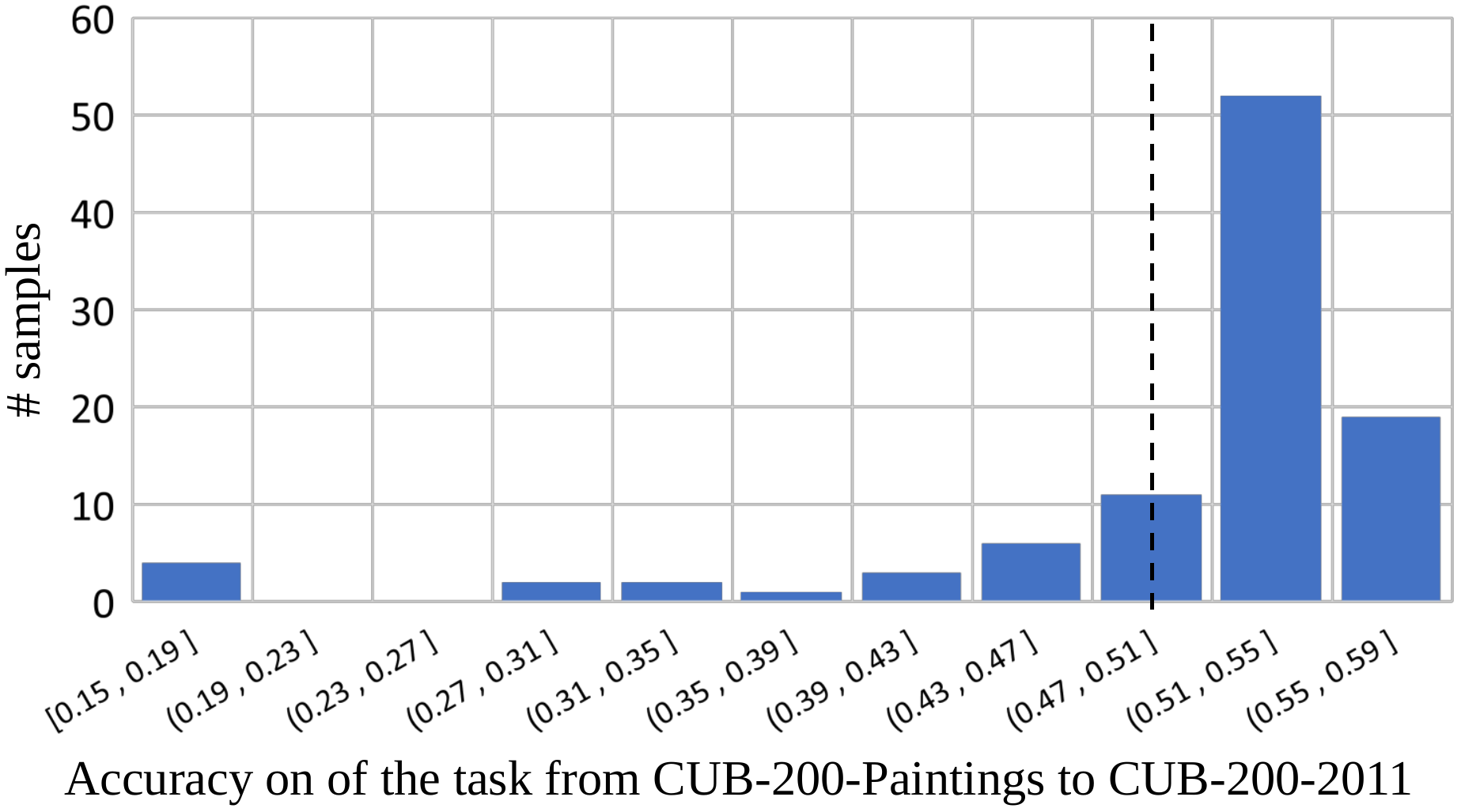}
        \end{minipage}
    }
    \vfill
    \subfigure[]{
        \label{fig:ArchAndTargetAcc}
        \begin{minipage}{\linewidth}
        \includegraphics[width=1\linewidth]{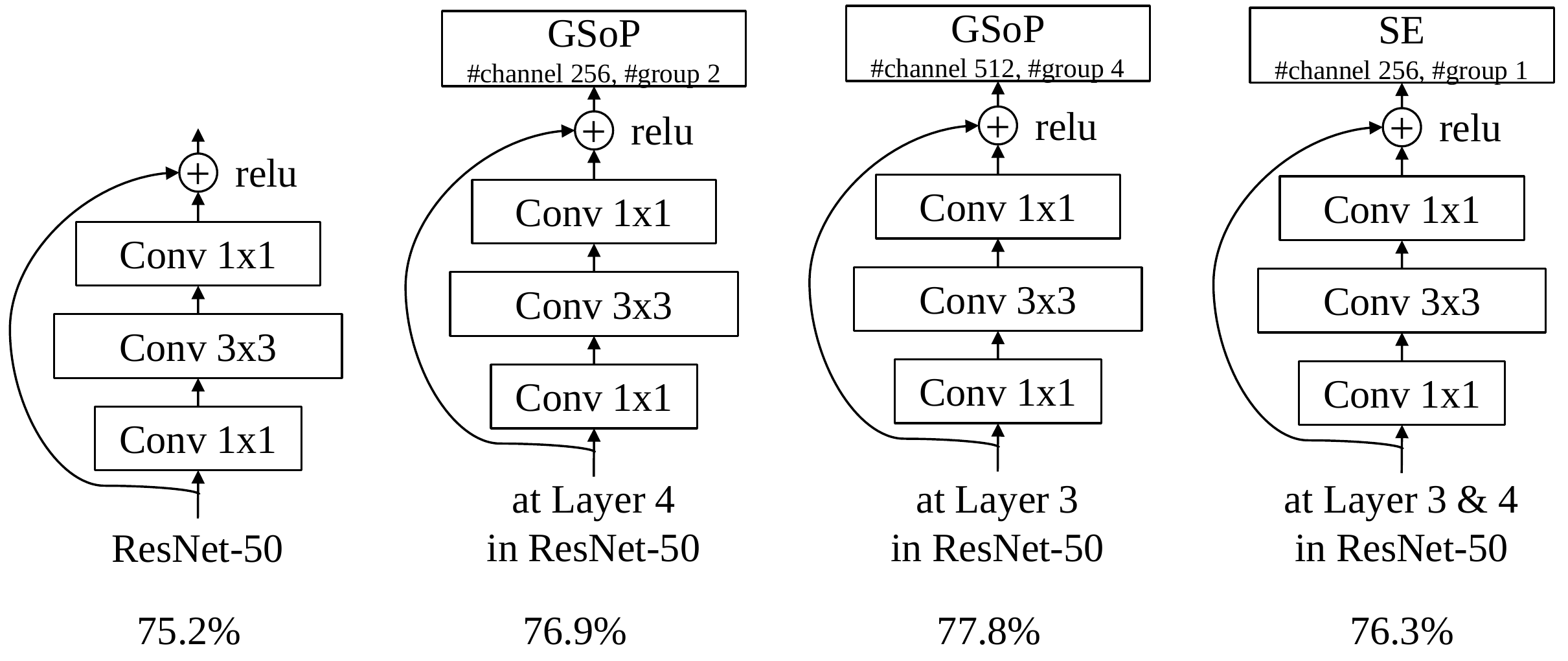}
        \end{minipage}
    }
    \caption{(a) Histogram of the accuracies for 500 random populations on the FGDA task of CUB-200-Painting to CUB-200-2011. The dashed vertical line indicates the result of ResNet-50.
    (b) Some seeds on the PDA task of Rw $\to$ Ar on Office-Home. The numbers indicate their corresponding accuracies on target domain.}
    \label{fig:HistOf500RandConfigs}
    \vspace{-4mm}
\end{figure}
\begin{figure}
    \centering
    \includegraphics[width=0.98\linewidth]{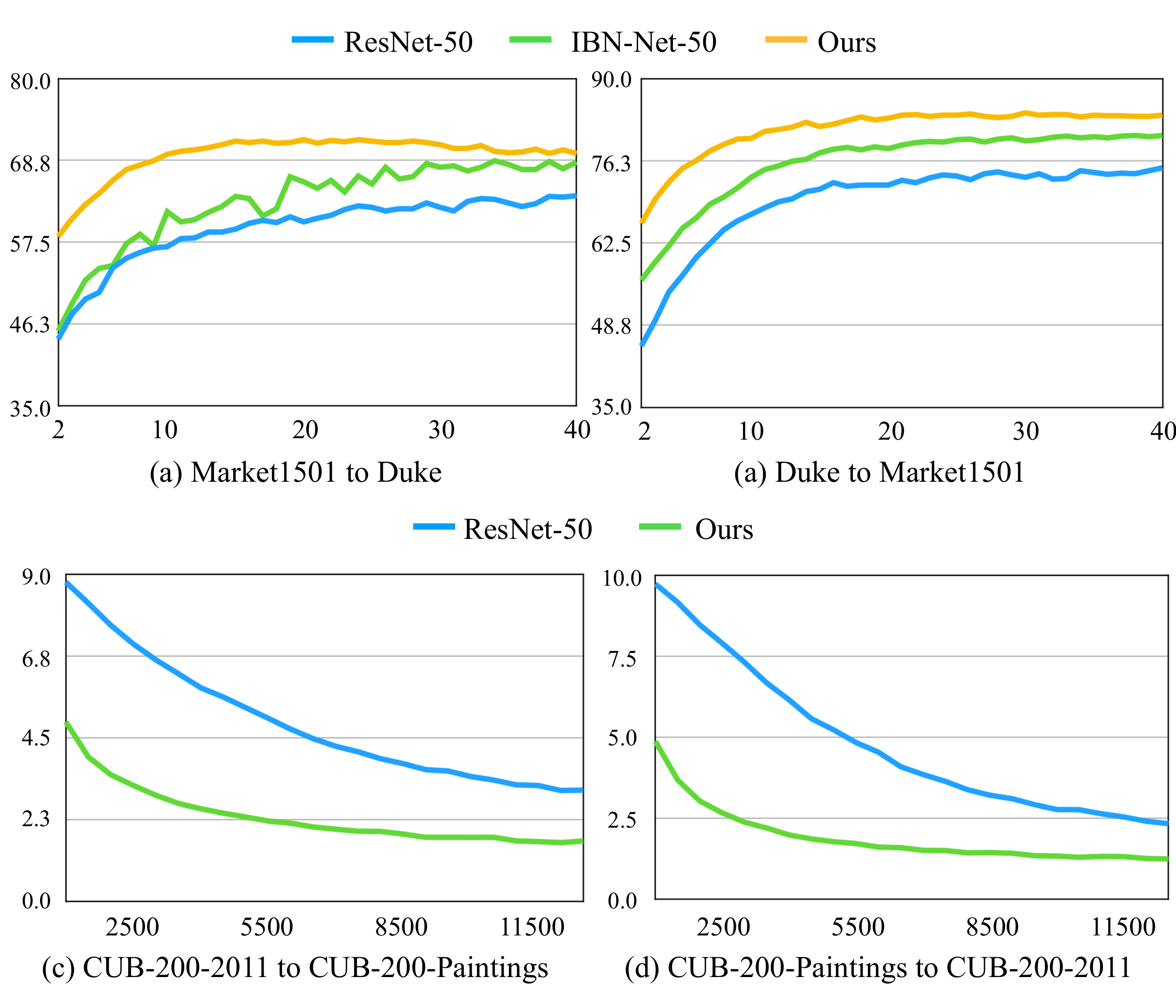}
    \caption{(a) and (b): The curves of three backbone networks on UDA tasks over person re-ID benchmarks. The x-axis is training epochs and the y-axis is the accuracy ($\%$) on target domain.
    (c) and (d): The differences between ResNet-50 and ours on fine-grained UDA tasks over CUB-Paintings. The x-axis indicates training iterations. The y-axis indicates the training loss.}
    \label{fig:VisTrainingEvalCurves}
    \vspace{-4mm}
\end{figure}

\vspace{-2mm}
\paragraph{Good and Bad Case Analysis.}
Finally, we take a closer look at the searched attention configurations. Figure~\ref{fig:HistOf500RandConfigs} displays the accuracies of $500$ randomly sampled populations on the FGDA task of CUB-200-Painting $\to$ CUB-200-2011.
The histogram verifies the benefit from refining the attention configurations and the effectiveness of the proposed attention configuration.
For better understanding the searched optimal architectures, we also visualize some attention configurations with good and sub-optimal UDA results (Figure~\ref{fig:ArchAndTargetAcc}), and the training curves of the optimal networks (Figure~\ref{fig:VisTrainingEvalCurves}).
Experiments indicate that \texttt{Layer 3} and \texttt{Layer 4} seem to be optimal positions to introduce attention modules, and we achieve the gains in accuracy when only moderate amounts of parameters and \#FLOPs are introduced.
All these numerical results can help and encourage researchers to cast a new light on designing novel attention modules towards better domain adaptation.

\section{Conclusion}
In this paper, we devise a novel and effective NAS algorithm for UDA problems.
We propose a more generalized way to apply the attention module for domain adaptation: to automatically optimize the attention configuration for one arbitrary UDA dataset.
We propose a new search space with a set of attention modules and their positions in the backbone network.
To be consonant with UDA settings, we propose a UDA-oriented estimation strategy: train the weights on two domains and evaluate the attention configurations in the target domain with a self-training pseudo-label strategy.
We implement the \iMethod\ framework based on an evolution-based NAS algorithm. 
Extensive experiments on multiple cross-domain benchmarks and typical adaptation scenarios verify that our scheme generally promotes popular domain adaptation methods.

For future work, we will investigate the transferability of various architectures and study the topic in other scenarios, \eg, object detection and semantic segmentation.

{\small
\bibliographystyle{ieee_fullname}
\bibliography{MAIN}

\begin{thebibliography}{10}\itemsep=-1pt

\bibitem{back1996evolutionary}
Thomas Back.
\newblock {\em Evolutionary algorithms in theory and practice: evolution
  strategies, evolutionary programming, genetic algorithms}.
\newblock Oxford university press, 1996.

\bibitem{bender2018oneshotNAS}
Gabriel Bender, Pieter-Jan Kindermans, Barret Zoph, Vijay Vasudevan, and Quoc
  Le.
\newblock Understanding and simplifying one-shot architecture search.
\newblock In {\em ICML}, 2018.

\bibitem{cai2018PathLevelNAS}
Han Cai, Jiacheng Yang, Weinan Zhang, Song Han, and Yong Yu.
\newblock Path-level network transformation for efficient architecture search.
\newblock In {\em ICML}, 2018.

\bibitem{cao2018SAN}
Zhangjie Cao, Mingsheng Long, Jianmin Wang, and Michael~I Jordan.
\newblock Partial transfer learning with selective adversarial networks.
\newblock In {\em CVPR}, 2018.

\bibitem{cao2019ETN}
Zhangjie Cao, Kaichao You, Mingsheng Long, Jianmin Wang, and Qiang Yang.
\newblock Learning to transfer examples for partial domain adaptation.
\newblock In {\em CVPR}, 2019.

\bibitem{caron2018DeepCluster}
Mathilde Caron, Piotr Bojanowski, Armand Joulin, and Matthijs Douze.
\newblock Deep clustering for unsupervised learning of visual features.
\newblock In {\em ECCV}, 2018.

\bibitem{chen2019BSP}
Xinyang Chen, Sinan Wang, Mingsheng Long, and Jianmin Wang.
\newblock Transferability vs. discriminability: Batch spectral penalization for
  adversarial domain adaptation.
\newblock In {\em ICML}, 2019.

\bibitem{chen2019progressive}
Xin Chen, Lingxi Xie, Jun Wu, and Qi Tian.
\newblock Progressive differentiable architecture search: Bridging the depth
  gap between search and evaluation.
\newblock In {\em ICCV}, 2019.

\bibitem{chen2019detnas}
Yukang Chen, Tong Yang, Xiangyu Zhang, Gaofeng Meng, Xinyu Xiao, and Jian Sun.
\newblock Detnas: Backbone search for object detection.
\newblock In {\em NeurIPS}, 2019.

\bibitem{chen2021DGFaceAntiSpoofing}
Zhihong Chen, Taiping Yao, Kekai Sheng, Shouhong Ding, Ying Tai, Jilin Li,
  Feiyue Huang, and Xinyu Jin.
\newblock Generalized representation learning for mixture domain face
  anti-spoofing.
\newblock In {\em AAAI}, 2021.

\bibitem{cui2020HDA}
Shuhao Cui, Xuan Jin, Shuhui Wang, Yuan He, and Qingming Huang.
\newblock Heuristic domain adaptation.
\newblock In {\em NeurIPS}, 2020.

\bibitem{cui2020BNM}
Shuhao Cui, Shuhui Wang, Junbao Zhuo, Liang Li, Qingming Huang, and Qi Tian.
\newblock Towards discriminability and diversity: Batch nuclear-norm
  maximization under label insufficient situations.
\newblock In {\em CVPR}, 2020.

\bibitem{elsken2019NASASurvey}
Thomas Elsken, Jan~Hendrik Metzen, Frank Hutter, et~al.
\newblock Neural architecture search: A survey.
\newblock {\em JMLR}, 2019.

\bibitem{falkner2018bohb}
Stefan Falkner, Aaron Klein, and Frank Hutter.
\newblock Bohb: Robust and efficient hyperparameter optimization at scale.
\newblock In {\em ICML}, 2018.

\bibitem{ganin2015DANN}
Yaroslav Ganin, Evgeniya Ustinova, Hana Ajakan, Pascal Germain, Hugo
  Larochelle, Fran{\c{c}}ois Laviolette, Mario Marchand, and Victor Lempitsky.
\newblock Domain-adversarial training of neural networks.
\newblock In {\em JMLR}, 2016.

\bibitem{gao2019GSoP}
Zilin Gao, Jiangtao Xie, Qilong Wang, and Peihua Li.
\newblock Global second-order pooling convolutional networks.
\newblock In {\em CVPR}, 2019.

\bibitem{ge2020MMT}
Yixiao Ge, Dapeng Chen, and Hongsheng Li.
\newblock Mutual mean-teaching: Pseudo label refinery for unsupervised domain
  adaptation on person re-identification.
\newblock In {\em ICLR}, 2020.

\bibitem{gomes2010discriminative}
Ryan Gomes, Andreas Krause, and Pietro Perona.
\newblock Discriminative clustering by regularized information maximization.
\newblock In {\em NeurIPS}, 2010.

\bibitem{guo2020SPOS}
Zichao Guo, Xiangyu Zhang, Haoyuan Mu, Wen Heng, Zechun Liu, Yichen Wei, and
  Jian Sun.
\newblock Single path one-shot neural architecture search with uniform
  sampling.
\newblock In {\em ECCV}, 2020.

\bibitem{he2016resnet}
Kaiming He, Xiangyu Zhang, Shaoqing Ren, and Jian Sun.
\newblock Deep residual learning for image recognition.
\newblock In {\em CVPR}, 2016.

\bibitem{hu2018SENet}
Jie Hu, Li Shen, and Gang Sun.
\newblock Squeeze-and-excitation networks.
\newblock In {\em CVPR}, 2018.

\bibitem{hu2017learning}
Weihua Hu, Takeru Miyato, Seiya Tokui, Eiichi Matsumoto, and Masashi Sugiyama.
\newblock Learning discrete representations via information maximizing
  self-augmented training.
\newblock In {\em ICML}, 2017.

\bibitem{kang2019CAN}
Guoliang Kang, Lu Jiang, Yi Yang, and Alexander~G Hauptmann.
\newblock Contrastive adaptation network for unsupervised domain adaptation.
\newblock In {\em CVPR}, 2019.

\bibitem{kurmi2019CADA}
Vinod~Kumar Kurmi, Shanu Kumar, and Vinay~P Namboodiri.
\newblock Attending to discriminative certainty for domain adaptation.
\newblock In {\em CVPR}, 2019.

\bibitem{li2020DCAN}
Shuang Li, Chi~Harold Liu, Qiuxia Lin, Binhui Xie, Zhengming Ding, Gao Huang,
  and Jian Tang.
\newblock Domain conditioned adaptation network.
\newblock In {\em AAAI}, 2020.

\bibitem{li2020adapting}
Yanxi Li, Zhaohui Yang, Yunhe Wang, and Chang Xu.
\newblock Adapting neural architectures between domains.
\newblock In {\em NeurIPS}, 2020.

\bibitem{liang2020shot}
Jian Liang, Dapeng Hu, and Jiashi Feng.
\newblock Do we really need to access the source data? source hypothesis
  transfer for unsupervised domain adaptation.
\newblock In {\em ICML}, 2020.

\bibitem{liang2020BA3US}
Jian Liang, Yunbo Wang, Dapeng Hu, Ran He, and Jiashi Feng.
\newblock A balanced and uncertainty-aware approach for partial domain
  adaptation.
\newblock In {\em ECCV}, 2020.

\bibitem{liu2019AutoDeepLab}
Chenxi Liu, Liang-Chieh Chen, Florian Schroff, Hartwig Adam, Wei Hua, Alan~L
  Yuille, and Li Fei-Fei.
\newblock Auto-deeplab: Hierarchical neural architecture search for semantic
  image segmentation.
\newblock In {\em CVPR}, 2019.

\bibitem{liu2019STA}
Hong Liu, Zhangjie Cao, Mingsheng Long, Jianmin Wang, and Qiang Yang.
\newblock Separate to adapt: Open set domain adaptation via progressive
  separation.
\newblock In {\em CVPR}, 2019.

\bibitem{liu2018DARTS}
Hanxiao Liu, Karen Simonyan, and Yiming Yang.
\newblock Darts: Differentiable architecture search.
\newblock In {\em ICLR}, 2018.

\bibitem{long2018CDAN}
Mingsheng Long, Zhangjie Cao, Jianmin Wang, and Michael~I Jordan.
\newblock Conditional adversarial domain adaptation.
\newblock In {\em NeurIPS}, 2018.

\bibitem{long2017JAN}
Mingsheng Long, Han Zhu, Jianmin Wang, and Michael~I Jordan.
\newblock Deep transfer learning with joint adaptation networks.
\newblock In {\em ICML}, 2017.

\bibitem{pan2018IBNNet}
Xingang Pan, Ping Luo, Jianping Shi, and Xiaoou Tang.
\newblock Two at once: Enhancing learning and generalization capacities via
  ibn-net.
\newblock In {\em ECCV}, 2018.

\bibitem{paszke2019pytorch}
Adam Paszke, Sam Gross, Francisco Massa, Adam Lerer, James Bradbury, Gregory
  Chanan, Trevor Killeen, Zeming Lin, Natalia Gimelshein, Luca Antiga, et~al.
\newblock Pytorch: An imperative style, high-performance deep learning library.
\newblock In {\em NeurIPS}, 2019.

\bibitem{quan2019AutoReID}
Ruijie Quan, Xuanyi Dong, Yu Wu, Linchao Zhu, and Yi Yang.
\newblock Auto-reid: Searching for a part-aware convnet for person
  re-identification.
\newblock In {\em ICCV}, 2019.

\bibitem{quionero2009dataset}
Joaquin Quionero-Candela, Masashi Sugiyama, Anton Schwaighofer, and Neil~D
  Lawrence.
\newblock {\em Dataset shift in machine learning}.
\newblock The MIT Press, 2009.

\bibitem{real2019EvoNAS}
Esteban Real, Alok Aggarwal, Yanping Huang, and Quoc~V Le.
\newblock Regularized evolution for image classifier architecture search.
\newblock In {\em AAAI}, 2019.

\bibitem{ristani2016Duke}
Ergys Ristani, Francesco Solera, Roger Zou, Rita Cucchiara, and Carlo Tomasi.
\newblock Performance measures and a data set for multi-target, multi-camera
  tracking.
\newblock In {\em ECCV}, 2016.

\bibitem{robbiano2021adversarial}
Luca Robbiano, Muhammad Rameez~Ur Rahman, Fabio Galasso, Barbara Caputo, and
  Fabio~Maria Carlucci.
\newblock Adversarial branch architecture search for unsupervised domain
  adaptation.
\newblock {\em arXiv preprint arXiv:2102.06679}, 2021.

\bibitem{saenko2010office31}
Kate Saenko, Brian Kulis, Mario Fritz, and Trevor Darrell.
\newblock Adapting visual category models to new domains.
\newblock In {\em ECCV}, 2010.

\bibitem{saito2020DANCE}
Kuniaki Saito, Donghyun Kim, Stan Sclaroff, and Kate Saenko.
\newblock Universal domain adaptation through self supervision.
\newblock In {\em NeurIPS}, 2020.

\bibitem{saito2018MCD}
Kuniaki Saito, Kohei Watanabe, Yoshitaka Ushiku, and Tatsuya Harada.
\newblock Maximum classifier discrepancy for unsupervised domain adaptation.
\newblock In {\em CVPR}, 2018.

\bibitem{saito2018OSBP}
Kuniaki Saito, Shohei Yamamoto, Yoshitaka Ushiku, and Tatsuya Harada.
\newblock Open set domain adaptation by backpropagation.
\newblock In {\em ECCV}, 2018.

\bibitem{venkateswara2017officehome}
Hemanth Venkateswara, Jose Eusebio, Shayok Chakraborty, and Sethuraman
  Panchanathan.
\newblock Deep hashing network for unsupervised domain adaptation.
\newblock In {\em CVPR}, 2017.

\bibitem{wah2011CUB-200-2011}
Catherine Wah, Steve Branson, Peter Welinder, Pietro Perona, and Serge
  Belongie.
\newblock The caltech-ucsd birds-200-2011 dataset.
\newblock In {\em California Institute of Technology}, 2011.

\bibitem{wang2020PAN}
Sinan Wang, Xinyang Chen, Yunbo Wang, Mingsheng Long, and Jianmin Wang.
\newblock Progressive adversarial networks for fine-grained domain adaptation.
\newblock In {\em CVPR}, 2020.

\bibitem{wang2019TN}
Ximei Wang, Ying Jin, Mingsheng Long, Jianmin Wang, and Michael~I Jordan.
\newblock Transferable normalization: Towards improving transferability of deep
  neural networks.
\newblock In {\em NeurIPS}, 2019.

\bibitem{wang2019fully}
Xijun Wang, Meina Kan, Shiguang Shan, and Xilin Chen.
\newblock Fully learnable group convolution for acceleration of deep neural
  networks.
\newblock In {\em CVPR}, 2019.

\bibitem{wang2019TADA}
Ximei Wang, Liang Li, Weirui Ye, Mingsheng Long, and Jianmin Wang.
\newblock Transferable attention for domain adaptation.
\newblock In {\em AAAI}, 2019.

\bibitem{wang2020DAFD}
Ze Wang, Xiuyuan Cheng, Guillermo Sapiro, and Qiang Qiu.
\newblock A dictionary approach to domain-invariant learning in deep networks.
\newblock In {\em NeurIPS}, 2020.

\bibitem{woo2018cbam}
Sanghyun Woo, Jongchan Park, Joon-Young Lee, and In So~Kweon.
\newblock Cbam: Convolutional block attention module.
\newblock In {\em ECCV}, 2018.

\bibitem{xu2019SAFN}
Ruijia Xu, Guanbin Li, Jihan Yang, and Liang Lin.
\newblock Larger norm more transferable: An adaptive feature norm approach for
  unsupervised domain adaptation.
\newblock In {\em ICCV}, 2019.

\bibitem{yosinski2014transferable}
Jason Yosinski, Jeff Clune, Yoshua Bengio, and Hod Lipson.
\newblock How transferable are features in deep neural networks?
\newblock In {\em NeurIPS}, 2014.

\bibitem{you2019UAN}
Kaichao You, Mingsheng Long, Zhangjie Cao, Jianmin Wang, and Michael~I Jordan.
\newblock Universal domain adaptation.
\newblock In {\em CVPR}, 2019.

\bibitem{zheng2015Market1501}
Liang Zheng, Liyue Shen, Lu Tian, Shengjin Wang, Jingdong Wang, and Qi Tian.
\newblock Scalable person re-identification: A benchmark.
\newblock In {\em ICCV}, 2015.

\bibitem{zoph2018NASNet}
Barret Zoph, Vijay Vasudevan, Jonathon Shlens, and Quoc~V Le.
\newblock Learning transferable architectures for scalable image recognition.
\newblock In {\em CVPR}, 2018.

\bibitem{zou2019CRST}
Yang Zou, Zhiding Yu, Xiaofeng Liu, BVK Kumar, and Jinsong Wang.
\newblock Confidence regularized self-training.
\newblock In {\em ICCV}, 2019.

\end{thebibliography}
}

\end{document}